\documentclass{article}
\pdfoutput=1

\usepackage[final]{ap_2020}

\usepackage[utf8]{inputenc} 
\usepackage[T1]{fontenc}    
\usepackage{hyperref}       
\usepackage{url}            
\usepackage{booktabs}       
\usepackage{amsfonts}       
\usepackage{nicefrac}       
\usepackage{microtype}      
\usepackage{float}
\usepackage{graphicx}
\usepackage{subcaption}
\usepackage{caption}
\DeclareGraphicsExtensions{.png,.pdf}
\usepackage[normalem]{ulem}

\newcommand{\Sref}[1]{\S\ref{#1}}

\usepackage{siunitx}

\usepackage[table]{xcolor}

\usepackage{textcomp}

\usepackage{makecell}

\usepackage{xspace}

\title{Audrey: A Personalized Open-Domain Conversational Bot}

\author{
Chung Hoon Hong\textsuperscript{2}, Yuan Liang\textsuperscript{2},  Sagnik Sinha Roy\textsuperscript{1}, Arushi Jain\textsuperscript{1}, Vihang Agarwal\textsuperscript{3} \\
\textbf{Ryan Draves}\textsuperscript{3}, \textbf{Zhizhuo Zhou}\textsuperscript{3}, \textbf{William Chen}\textsuperscript{3},  \textbf{Yujian Liu}\textsuperscript{3}, \textbf{Martha Miracky}\textsuperscript{3}, \textbf{Lily Ge}\textsuperscript{3}\\
\textbf{Nikola Banovic}\textsuperscript{3}, \textbf{David Jurgens} \textsuperscript{1,3}\\
\textsuperscript{1}School of Information\\
\textsuperscript{2}Department of Statistics\\
\textsuperscript{3}Department of Electrical Engineering \& Computer Science\\
University of Michigan\\
\texttt{\{datahong, yualiang, sagniksr, arushij, vihang, dravesr, zhizhuo\}@umich.edu} \\
\texttt{\{wxllchxn, yujianl, mmiracky, iterm, nbanovic, jurgens\}@umich.edu} \\
}

\begin{document}

\maketitle

\begin{abstract}



Conversational Intelligence requires that a person engage on informational, personal and relational levels. Advances in Natural Language Understanding have helped recent chatbots succeed at dialog on the informational level. However, current techniques still lag for conversing with humans on a personal level and fully relating to them. The University of Michigan's submission to the Alexa Prize Grand Challenge 3, Audrey, is an open-domain conversational chat-bot that aims to engage customers on these levels through interest driven conversations guided by customers' personalities and emotions. Audrey is built from socially-aware models such as Emotion Detection and a Personal Understanding Module to grasp a deeper understanding of users' interests and desires. Our architecture interacts with customers using a hybrid approach balanced between knowledge-driven response generators and context-driven neural response generators to cater to all three levels of conversations. During the semi-finals period, we achieved an average cumulative rating of 3.25 on a 1-5 Likert scale.       

\end{abstract}

\section{Introduction}

Naturally conversing with artificial agents has been a lofty goal since the beginning of the computing era, starting with the Turing Test. The tremendous growth in the Conversational AI paradigm in the recent decade has brought conversational agents---chatbots---closer to this goal, as the research community has become increasingly interested in systematically developing and testing these models. Goal-oriented chatbots have seen significant growth and adoption in areas such as basic question answering services online \cite{brandtzaeg2017people}. The success of goal-oriented chatbots lies in their ability to carry out a meaningful and useful conversation in a limited domain where the range of topics and user utterances is restricted and predictable (e.g., booking a plane ticket or offering limited helpdesk advice). Yet, open domain chatbots face substantial challenges in having similar levels of success, as these need to understand diverse context from potentially any domain, determine how to respond to such content in a way that makes for a natural conversation (beyond just the response level), and generate human-like responses.

We took a step towards the vision of naturally conversing artificial agents and built Audrey, an open domain chatbot that tackles all of the main challenges posed to open domain chatbots.  Audrey participated in the Alexa Prize Socialbot Grand Challenge 3 which provided us a platform to implement and deploy Audrey to a broad audience. Audrey was first deployed to Amazon Alexa customers on December, 4\textsuperscript{th}, 2019 and this report summarizes our conversations with Alexa customers until the end of semi-finals interaction period on April 29\textsuperscript{th}, 2020. When the customer invokes ``let’s chat,'' Audrey was randomly chosen from one of the ten Alexa Prize socialbots for interaction.

We constructed Audrey using multiple technical models in three thematically-grouped components: (1) natural language understanding, (2) dialog management, and (3) response generation. 
The first of these components aims to understand what the customer has said at the semantic and social levels and includes models for tasks such as (i) Noun Phrase Extraction, (ii) Sentiment Classification, and (iii) Emotion Classification. 
The second of these components aims to decide how to respond to the customer's speech based on goals for longer conversation. Here, we introduce multiple innovative models for handling this conversation policy, including (i) a Personality Understanding Module that infers the interests of customers, (ii) reinforcement learning for selecting conversation topics and (iii) an adaptive strategy for transitioning between template-based and neural-network-based response generators to maximize conversational coherence.
The third component encompasses a variety of modules for generating engaging responses to customers using different strategies including template-based generators, neural response generators, and hybrid generators with a mix of both for a seamless conversation flow. When used alone template based responses are often generic and fail to display all aspects of human-like attributes in conversations while neural response generators have difficulty tracking long-term aspects of the conversation. We propose to use the hybrid generator to deal with these shortcomings of either approach. 


%

Based on interactions with customers, our work offers the following three contributions towards the development of open domain chatbots. First, open domain chatbots tend to heavily gravitate towards either a rule-based system or end-to-end neural network approaches. Our work informs that such chatbots can benefit and realize new avenues for improvement by finding the right balance between these systems. Second, customers enjoy engaging with a chatbot on day-to-day topics such as fitness, pets, and technology. Additionally, they desire such bots to engage in more open-ended personal chit-chat. Designing improved modules that allow for quick discovery of customer preferences and intents (such as our model for inferring customers' interests in Section \ref{sec:pum}) will enable open domain chatbots to be deployed as naturally conversing artificial companions right from the start of the conversation. Finally, leveraging context information that accommodates for various customer behaviors and designing a mix of dialogue policies for making high-level decisions allows coherent dialogues and a smooth dialogue-flow in open domain chatbots.  

We evaluated Audrey by analyzing the impact of generators and conversation starters in Section \ref{sec:analysis}. We also conducted experiments on our generators and dialogue policy to support our claims in Section \ref{sec:experiments}. 





\section{Architecture}
\begin{figure}[hbt!]
    \centering
        \includegraphics[width=1.0\textwidth]{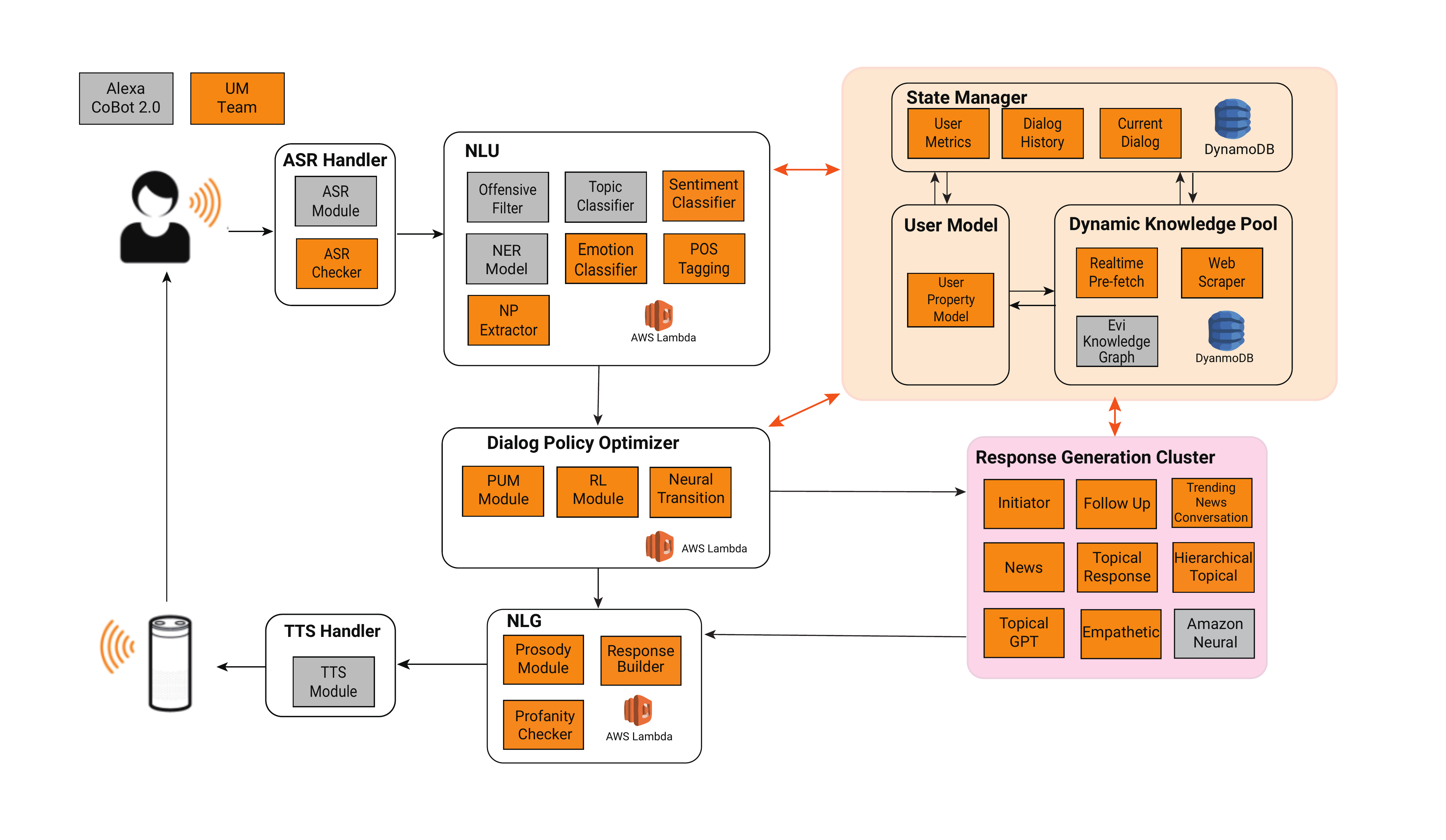}
        \caption{Audrey Architecture}
        \label{fig:architecture}
    \end{figure}

We designed Audrey as a modular, scalable framework that allows rapid iterative testing and high-availability deployment. Audrey is implemented on top of the Amazon Conversational Bot Toolkit (CoBot) \cite{ch2018advancing}. The core concepts behind our architecture, such as noun phrase extraction, emotion classification, and response generators, are deployed as APIs on independent Docker modules. A server-less AWS lambda function is used to interface with Alexa Skills Kit (ASK) and connect our Docker modules, creating the Audrey architecture. Figure \ref{fig:architecture} shows our modular framework and how the conversation state is processed and updated during the conversation.

Audrey's system consists of three core components:
\begin{itemize}
    \item Natural Language Understanding (NLU), which processes input from the customers (\Sref{sec:nlu})
    \item Dialog Policy Optimizer, which recommends the most relevant conversation topic and provides natural transitions between topics (\Sref{sec:dp})
    \item Natural Language Generation (NLG), which handles different facets of social conversations through a mix of template-based generators, topic-based retrieval generators, and neural response generators for handling out of domain conversations (\Sref{sec:nlg})
\end{itemize}

\subsection{Core Component 1: Natural Language Understanding} \label{sec:nlu}

In a social dialog, it is important to discuss subjects that are relevant to the conversation and to understand the nuances of the conversation. The NLU modules focus on finding and extracting the most relevant information to our template based generators using noun phrase extraction and Amazon entity recognition. The NLU component also provides context information for transitions to dialogue policy using an emotion classifier and a sentiment classifier to inform how customers reacted to the previous turn.


\subsubsection{Noun Phrase Extraction}

Noun phrases in a conversation help identify the topics and other important information from the speaker. Through the noun phrases we get a better understanding of the topic, which helps in activating the appropriate topical module. We first used a noun phrase extraction model based on Spacy \cite{honnibal-johnson:2015:EMNLP} to recognize key concepts that customers mentioned in the conversation. However, we found the extraction model's performance was low in our conversational setting, which hurt our ability to recognize key concepts the customers talked about.

Therefore, we chose to deploy a state-of--the-art model to deal with the noun phrase extraction task. Specifically, we use the Bidirectional Encoder Representations from Transformers model (BERT) \cite{devlin2018bert} as the backbone noun phrase extraction model. Although BERT can deal with many natural language understanding tasks, here we mainly leverage part of speech (POS) tagging as the downstream task of BERT. During the inference step, we set up the POS tag combination rule to extract the noun phrases. Compared to Name Entity Recognition (NER), the method we use is much more flexible because we can adjust the extracting rule to extract the noun phrases as per our interest.

We fine-tuned the pre-trained BERT model with PennTree bank dataset \cite{10.5555/972470.972475}. We compared the performance of our fine-tuned BERT model with Condition Random Field (CRF) and Bi-LSTM+CRF. The result in Table \ref{tab:POStagging} shows that BERT clearly outperforms the other two models.

\begin{table}[h]
  \centering
  \begin{tabular}{ll}
    \toprule
    \cmidrule{1-2}
    Models     & Accuracy(\%)    \\
    \midrule
    CRF & 77.63      \\
    Bi-LSTM + CRF & 89.57 \\
    BERT & 94.97 \\
    \bottomrule
  \end{tabular}
  \setlength{\abovecaptionskip}{3pt}
  \caption{Performance Comparison for POS Tagging}
  \label{tab:POStagging}
\end{table}

\subsubsection{Entity Resolution} \label{sec:kg}

Entity Resolution allows us to connect concepts mentioned by the customer to broader knowledge in order to continue a conversation along the topic. We utilize the Entity Resolution service from Amazon Evi Knowledge Graph to find relevant entities related to the extracted noun. For example, suppose our Noun Phrase Extraction extracted the noun phrase ``Avatar'' in previous step; the Entity Resolution service would recognize the noun phrase as the entity \textit{movie:Avatar}.  We would then use predefined custom queries  for movie related topics for Amazon Evi Knowledge Graph to find related director \textit{director:James Cameron} and actor \textit{Sam Worthington}.  We would then pass the extracted information to the hierarchical generator described in Section \ref{sec:hiearchy}.

\subsubsection{Sentiment Classification}\label{sec:sentiment}

Sentiment Classification is a way to computationally classify text into positive, negative or neutral opinions. It implicitly allows Audrey to understand customers and their preferences while providing a way to make informed decisions about the conversation's trajectory. We used a lexicon and rule-based sentiment classifier called Vader \cite{vader} to assign a sentiment score to input customer utterances. We maintained a global sentiment for each conversation by calculating the running average of sentiment scores. These sentiments were principally used in strategy selection and topical transitions which we describe in detail in transitions Section \ref{sec:dp} under dialogue policy.

\subsubsection{Emotion Classification}

\begin{figure}[hbt!]
\centering
    \includegraphics[width=0.6\textwidth]{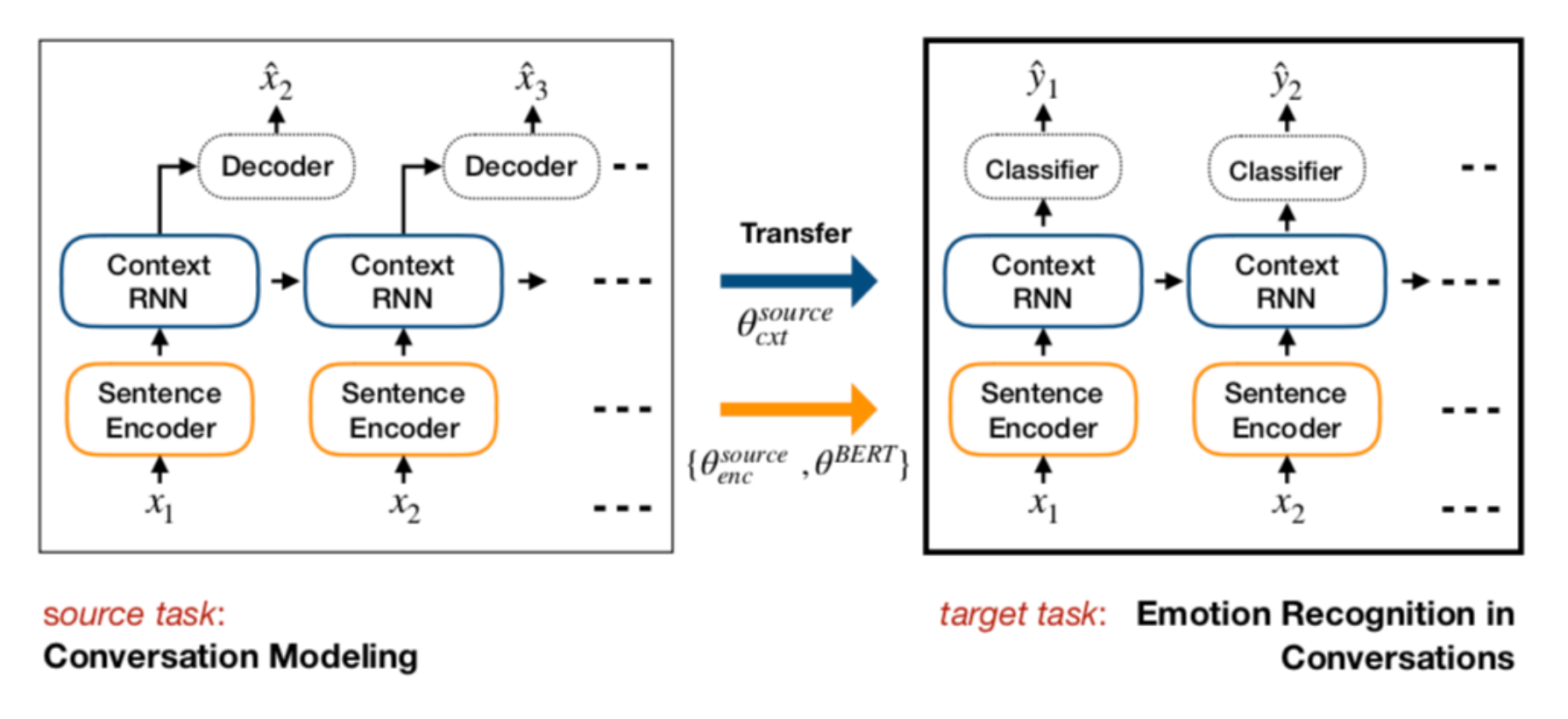}
    \caption{Framework for Emotion Recognition in Conversations (ERC) using Transfer Learning (TL) parameters. The knowledge from a dialogue generator is transferred into the discriminative task of emotion recognition.}
    \label{fig:ERC}
\end{figure}

One of the challenges of any dialogue agent is recognizing the feelings of the conversation partner and replying accordingly. Customers feel more satisfied when given a response that is generated by understanding their underlying emotions. In order to give an empathetic response, understanding the underlying emotion of the conversation is of great importance. We utilize a emotion detection model named TL-ERC \cite{hazarika2019emotion} to deal with emotion classification task (Figure \ref{fig:ERC}). TL-ERC \cite{hazarika2019emotion} is a two-stage model in which the first part is a generative conversation model constructed by a sentence encoder, a context encoder, and a sentence decoder, while the second part is a emotion recognition model containing a sentence encoder, a context encoder, and a classifier.  

The sentence encoder of emotion recognition model is fine-tuned from the pretrained checkpoint from BERT, and the context encoder from the corresponding component of the generative conversation model. For pretraining with the source task of the conversation model, we use Cornell Movie Dialog Corpus \cite{Danescu-Niculescu-Mizil+Lee:11a}, a large scale benchmark dataset. Here we compare the performance between TL-ERC and Fasttext\cite{joulin2016bag}. The result is shown in Table \ref{tab:emotion}, from which we can see that TL-ERC is more powerful. For fine-tuning emotion recognition model, we use Empathetic Dialogues\cite{rashkin2019towards}, a novel data set of 25,000 conversations grounded in emotional situations released by Facebook in 2019. In the original dataset, there are total 31 different emotion categories. However, there are some similar kinds of emotions in the dataset such as "joyful" \& "impressed", "annoyed" \& "furious", etc. To simplify further tasks, we grouped these 31 emotion labels into 10 emotion labels based on their meaning and similarity and formed a new dataset for training. 


\begin{table}[h!]
  \centering
  \begin{tabular}{lllll}
    \toprule
    \cmidrule{1-5}
    Models    &  Accuracy(\%) & Avg Precision(\%) & Avg Recall(\%) & Avg F1(\%)  \\
    \midrule
    Fasttext  & 52.99         & 53.51             & 52.99          & 53.10     \\
    TL-ERC    & 61.82         & 62.82             & 61.82          & 61.88    \\
    \bottomrule
  \end{tabular}
  \setlength{\abovecaptionskip}{2pt}
  \caption{Performance Comparison for Emotion Classification}
  \label{tab:emotion}
\end{table}
\subsection{Core Component 2: Dialogue Policy} 
\label{sec:dp}



Audrey is a dialog agent designed for both topical and open domain chit-chat. For customer satisfaction, such an agent must not only have deep personal conversations with the customer but also provide customers an opportunity to converse on a breadth of topics. There can be several topics that a customer may like to talk about including popular ones like movies, sports, animals, etc. or other topics such as arts, gaming, or even Pokemon. Enabling a dialog agent with domain knowledge and expertise to handle these various avenues makes it equally important to manage dialogue flow for human like conversations. Managing dialogue requires Audrey to have a dialogue policy that tracks its state, smoothly transition from one avenue to other and guide conversations to topics that customers may find engaging. A crucial component of Audrey is to decide what topic to talk about at each turn, which is determined by the dialogue policy. We use the dialogue policy to help guide Audrey's conversations to relevant template based or neural generation based responses. 

Our Dialog Policy Optimizer component recommends the most relevant conversation topic for the customer as well as provides natural transitions between topics for our conversations. We introduced the Personality Understanding Module (PUM) that collects and stores customer information to infer their interests. We then leverage knowledge about the customers to select and recommend the most relevant, personalized topics to them using a Reinforcement Learning-based approach. To accommodate various customer behaviors and keep conversations coherent, our transition mechanisms guide the flow between topical and out-of-domain conversations. The transition mechanism uses sentiment classification and threshold based transitions for our neural generators.

\subsubsection{Topical Transitions}


A conversational agent that aims to engage and entertain customers requires the ability to maintain coherent conversations. The agent can maintain such coherent conversations by ensuring smooth topical transitions which play a very important role and are described as a dialogue policy below. Audrey's goal is to be a personal chatbot that can engage customers in topics they enjoy conversing about. Although Audrey can talk on a wide variety of topics, some customers found certain topics more engaging than the others. Additionally, we hypothesized that the customers would expect Audrey to guide the conversation without them explicitly mentioning what they wanted to talk about.

The overall goal of the dialogue policy is to maintain a positive global sentiment of each conversation while avoiding repeated negative sentiment customer-utterances. We use the sentiment classifier from Section \ref{sec:sentiment} to assign sentiment scores and maintain a global sentiment. The dialogue policy switches to a different topic when the sentiment score drops (we specified thresholds on sentiment scores to define such states), a strategy that we found to work well in practice.

Audrey decides on the new topic to transition to using one of two different approaches: (1) PUM (Section \ref{sec:pum}) or (2) an RL-based approach (Section \ref{sec:rl}). We observed smooth topic transitions can be ensured by asking questions to the customers. Thus, our topic transitions are always followed by a template based Initiator module (Section \ref{sec:initiators}) that ensures coherent dialogue flow. To handle out of topic transitions (e.g., generic chat and out of domain utterances), Audrey uses transitions based on Neural Response Generators (\ref{sec:transitions}). We will use the term \textit{transitions} throughout the rest of the paper to describe topical switches followed by a relevant generated question.

\subsubsection{Personal Understanding Module (PUM)} \label{sec:pum}


Chatbots that personalize a customer's experience could improve the quality of the experience, make the interaction easier, and make the customer feel understood.  To do this requires knowledge about customers that allows Audrey to pick topics that are relevant to the customer. We build the Personal Understanding Module (PUM) to offer personalized experience and direct the conversation based on different customer personalities. We only invoke the PUM module when the topic is exhausted or when we do not have enough context information to respond to the customer. 


When Audrey invokes the PUM, it asks customers a proxy question that could provide additional information about the customer's preference for different topics. For example, Audrey may ask customers their interests in books: \textit{"I am lucky to have access to every single book online. It's so easy to get lost in a good book. I personally like sci-fi and fantasy books. Do you like reading often?"} The answer to the proxy question allows us to set a customer attribute that corresponds to their answer (e.g., whether they like books, movies, sports, video games, etc.).

\begin{figure} [hbt!]
\centering
    \includegraphics[width=0.95\textwidth]{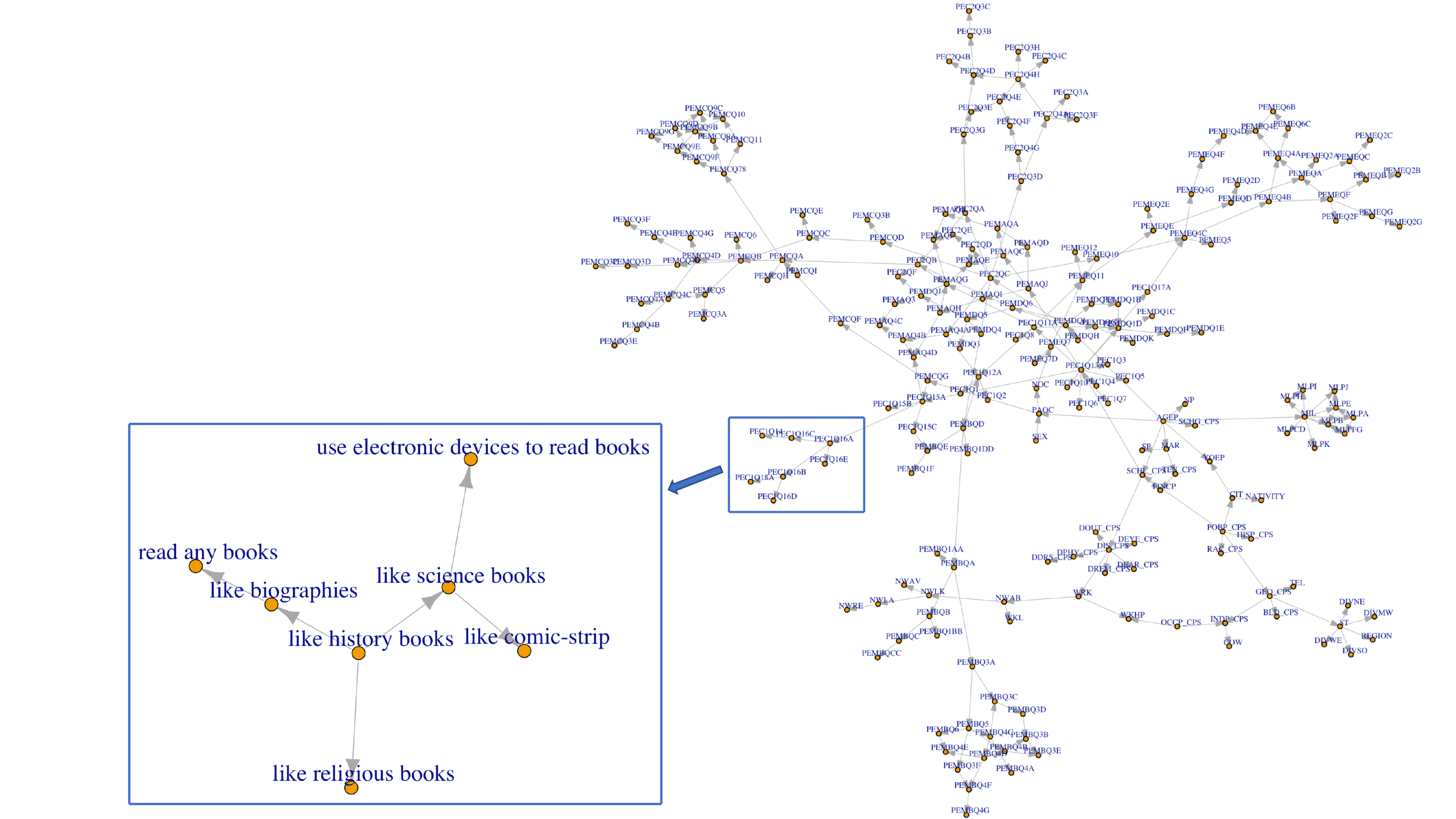}
    \caption{Bayesian network that models customer attributes. Each node in the network represents one attribute such as whether the customer likes books or not. The network allows Audrey to infer values of other customer attributes given knowledge about a subset of customer attributes.}
    \label{fig:pum}
\end{figure}

We have built a Bayesian network that models relationships between customer attributes (Figure \ref{fig:pum}). The network allows us to represent the conditional dependence between these attributes using a directed graph. We estimate the parameters of the network (i.e., conditional probabilities of different attributes) using data from the Survey of Public Participation in the Arts\cite{oftheCensus2015}.

Audrey uses the network to personalize the conversation and direct it to topics relevant to the customer. After each customer response to a proxy questions, Audrey can infer the probability of other attributes that it did not ask about based on the customer's previous answers. We relate each attribute to different topics Audrey can talk about to select the most likely topic of interest to the customer.

\subsubsection{Topic Selection based on Reinforcement Learning (RL)} \label{sec:rl}




To offer a truly personalized experience the chatbot needs to reason about and plan sequences of topics that the customer wants to talk about throughout the whole conversation. Audrey leverages Reinforcement Learning (RL) to estimate a policy that allows it to select the next topic based on customers' explicit preferences for certain topics and their ordering. When the topic is not exhausted, and the conversation with the customer is continuing, the RL based topic selection is used.


We formulated the problem of topic selection as a Markov Decision Process\cite{suttonrl2017}. The task of the agent is to chose the next topic for transition conditioned on the current topic of conversation. We assume each topic selection to be associated with a reward from an unknown distribution. Ideally, we want these rewards to accurately approximate customer preferences. We train an agent to estimate these reward distributions given the training data. Our training data consists of Audrey's conversations along with customer ratings collected throughout the semifinals stage. These ratings are used as the feedback signal to train our agent. The agent is trained using an offline batch policy where the agent learns only from the data collected in the past without directly interacting with the environment. 

We represent each conversation $C_i$ as a sequence of $t$ unique topics ($C_i = \{a_0, a_1, a_2, .... a_t\}$) with the given rating $R_i$. Let the reward at each turn of the conversation be given by an unknown function $r(\cdot)$ as $r(a_{t-1}, a_t)$. In an online setting, the total expected reward of the conversation is given by:

\begin{equation}
\label{eq:reward}
\mathbb{E}_{C_i}(R) = \sum_{i=1}^{t} \gamma^{t-i} \cdot r(a_{i-1}, a_i),
\end{equation}

where $\gamma$ is the discount factor. Estimating $r(\cdot)$ would allow us to select topics that better estimate customer intention at each conversation turn. We represent $r(\cdot)$ as a multi-layer perceptron neural network. Our objective is to minimize the mean squared error between the expected reward $\mathbb{E}_{C_i}(R)$ and the rating $R_i$ and learn $r(\cdot)$ from the training data. For improved convergence during training, we scale the ratings s.t. $R_i \in [0,1]$. We set $\gamma$ as 0.99 in our experiments.  

During deployment, we fixed the learned $r(\cdot)$. Topics are selected during transitions using an $\epsilon-greedy$ action selection strategy. We choose topics either with the maximum estimated reward or uniformly at random with probability $\epsilon$. Such a formulation allows Audrey to learn about the sequence of topics customers find engaging from the past conversations.

\subsubsection{Transitions based on Neural Response Generators} \label{sec:transitions}


An ideal chatbot would be able to talk at depth about a topic the customer is passionate about, but should also be able to transition between topics when the conversation becomes stale. One indicator of stale topics is the customer's switch to generic chat or other out of domain utterances.

To handle generic chat and out of domain utterances, we leverage the power of neural response generators. We developed a module to intelligently decide the number of previous turns to pass to the generator by looking at the previous states and finding out how long the customer talked about the current topic. This way, the model was able to generate sophisticated responses when Audrey covered a topic in depth, while being flexible enough to switch topics if the customer initiated the transition. The specifics of our neural generators will be described in detail in the next section \ref{sec:nlg}.



\subsection{Core Component 3: Natural Language Generation} \label{sec:nlg}



Recent advances in Natural Language Understanding (NLU) have helped spark recent interest in conversational AI. However, chatbots and voice assistant still generate responses in very fixed and robotic manner. In order to produce more diverse and personalised output, chatbots need to be able to automatically generate language adapted to the current context. To accomplish our goal of an open-domain chatbot, we developed a Natural Language Generation (NLG) system comprised of a variety of  template-based and neural generation models, which can be selected and adapted based on context. These generation modules work together to produce responses for different stages of conversation from small talk to discussions of customers' interests, from acknowledging customers emotions to giving opinions.   

Mirroring real-world behavior when striking up a conversation with a stranger, Audrey starts all conversations with an ice-breaker question invoked by our Initiators module (\Sref{sec:initiators}) to make a great first impression and form an immediate connection with the customer. Next, Audrey deepens the conversation by engaging the customer with one of our topic modules using a Follow Up Response Generator (\Sref{sec:followup}).
There are four different retrieval-based topical modules which differ in their architectures but have one goal: in-depth and coherent conversation on a particular topic at length. To better handle popular topics like Movies, Books and Music, we developed the Hierarchical Topical Generator (\Sref{sec:hiearchy}) that uses a hierarchical structure about subtopics within a topic and can navigate within and between these subtopics using Amazon's Evi Knowledge Graph. Additionally, to further our goal of building a chatbot based diversity, we curated topics of customers' interests (weather, season, arts, gaming, and Pokemon etc.) and developed a dynamic retrieval-based generator to discuss these (\Sref{sec:topical-response}). Finally, to keep Audrey grounded in the real world, we designed two modules to discuss news: (1) the Trending News Conversation Generator (\Sref{sec:dynamic}) talks about recent news by continuously pulling trending information from social media and (2) the News Response Generator (\Sref{sec:newsgenerator}) which sources the latest news from on articles from Washington Post for discussion.  

All of these modules are templated retrieval based systems but for open-domain chat we require more than such systems and hence we deployed 3 different neural generation modules. Our TopicalGPT Response Generator (\Sref{sec:topical-response}) handles all the random topics which could not be handled by retrieval systems. It is developed by fine-tuning the GPT-2  language generation model by OpenAI   \cite{radford2019language} on Topical-Chat dataset released by Amazon  \cite{Gopalakrishnan2019}. We also developed the Empathetic Response Generator (\Sref{sec:empathetic}) to connect with customers on an emotional level by responding when we detect a customer has replied with emotion. Lastly, we also use Amazon's Neural Response Generator (\Sref{sec:amazonneural}) based on the transfer learning approach by HuggingFace  \cite{wolf2019transfertransfo}.   

\subsubsection{Initiators Module} \label{sec:initiators}

We believe that first time conversations thrive on ice-breakers which can break the awkward silence and establish a conversational common ground. With this idea in mind, we developed the initiator module to generate human-level ice-breaker questions with follow-ups. Instead of simply asking customers or opening up with a "hey, how are you doing," we took the initiative by asking them intriguing questions. Audrey asks customers questions like "what is that one thing which you want to do today?", "how many hours do you spend on your computer each day?", and "if you were to write a book about your life, what would it be called?" To further improve the transition, the Initiator Module would be followed by transitions by the Follow Up Response Generator, described in Section \ref{sec:followup}.

To make the first few turns as delightful as possible, we hand-crafted templates with questions and follow-ups, three turns deep. The optimal response template is selected via a weighted score of keywords search and the mean-pooled sentence vector cosine similarity using SpaCy \cite{honnibal-johnson:2015:EMNLP} word embeddings based on the customers' utterance to our initial ice-breaker question.

\subsubsection{Follow Up Response Generator}\label{sec:followup}

Natural transitions from one topic to another is key to engagement in a social conversation.  Audrey's architecture uses a Follow Up Response Generator to handle transitions in between topics.

In order to make smooth transitions from the initial icebreakers leading to deeper conversations in different topic modules, we use the Follow Up Response Generator. We have custom transitions for ten topical modules.  Initially, customers were given options to choose from general topics such as movies, books, and music.  Looking at our conversations, we realized that giving customers too many choices sometimes lead conversations to a deadlock.  Rather than that, we try to have natural transitions from previous initiator topic to our customized topical modules. For example before transitioning to Fitness module, Audrey says "I've started doing cardio recently! Getting stuck in my little electric box isn't really good for my health." which gives a more natural path towards topical discussion. 

The main transition mechanism described for neural generators in Section \ref{sec:transitions} were supplemented by a section of engaging conversational questions, e.g., ``What's the smartest thing you've seen your pet do ?''
Recognizing that engaging customers on a conversational trajectory that leads to deeper discussion can create a strong bond  \cite{aron1992inclusion}, we developed a novel procedure to rank questions by intimacy to help better engage customers.
3.0M questions were scraped from Reddit and used to fine-tune a BERT language model. Then the BERT model was trained to predict each question's intimacy [-1,1] using a training set of 960 questions rated for intimacy and developed by us. The model attained Pearson's $r=0.77$ on a held out test set indicating it is close to meeting human judgment on intimacy. 
After filtering all questions to a narrow range appropriate for discussion and use in an Alexa Prize chatbot, we categorized the questions into low, medium, and high intimate question. Experiments in Section \ref{sec:exp-intimacy} test the effect of question ordering on conversation ratings.


\subsubsection{Trending News Conversation Generator}
\label{sec:dynamic}
The purpose of the Trending News Conversation Generator was to introduce recent news and content into Audrey's conversations. While it is easier to deploy template based or generated conversations related to seasonal events or longstanding news, our aim with the Trending News Conversation generation was to introduce fresh content and topics that a customer may have heard in the past few days.

In particular, due to the recent events of COVID-19 dominating the news cycle and reducing the availability of positive content and entertainment, we primarily pulled content by scraping Reddit's /r/UpliftingNews subreddit, as the majority of the content is news articles with a positive focus, without a specific domain.

For dynamic data and knowledge, Audrey used SoundingBoard's data pipeline in order to retrieve, filter, process, and upload daily Reddit content to DynamoDb \cite{fang-etal-2018-sounding}. Subreddits that were dynamically scraped included /r/TodayILearned, /r/movies, /r/UpliftingNews, /r/news, /r/worldnews, /r/science, /r/sports, and other topical subreddits in the same vein. SoundingBoard's extensive filtering process, which ensured a minimal amount of sensitive content was uploaded to DynamoDb, was applied to all dynamic data, as well as much of the static datasets and our sensitive-content filtering for Audrey.


As shown in Table \ref{tab:newsexample}, the initial prompts of the trending news generator use positive news headlines across a variety of domains (with the exception of having some basis on COVID-19). From here, the trending news generator utilizes the Amazon neural generator from Section \ref{sec:amazonneural} to generate further responses, feeding in filtered Reddit comments from the original thread into the knowledge input of the Amazon neural generator. The Table \ref{tab:newsexample} shows a longer exchange:

\begin{table}[h]
    \begin{tabular}{l|p{10cm}}
    \textbf{Conversation} & \\
    Audrey & Have you heard that john krasinski is throwing a virtual prom for the class of 2020? \\
     \cellcolor{gray!5}User &  \cellcolor{gray!5} Wow, that's very nice of him. \\
    Audrey & It sure is. He is a very talented guy. \\
     \cellcolor{gray!5}User &  \cellcolor{gray!5} Yeah. I'm a huge fan of his. \\
    Audrey & He is also a very smart person. \\
     & \\
    \end{tabular}
    \caption{Conversation example of the Trending News Conversation generator. To respect customers’ privacy, this is not a real customer conversation.}\
    \label{tab:newsexample}
\end{table}

The Trending News Conversation generator's results can be found in \Sref{sec:tncg-exp}. Overall, there was no significant impact on ratings, but the average conversation duration dramatically increased. It is difficult to understand the impact COVID-19 has on its performance, as one is unable to access how the customer feels about the topic versus a time period where the news cycle is less focused on one particularly sad topic. Regardless, the Trending News Conversation generator is able to generate responses to recent news items by using dynamic web scraping in combination with Amazon Neural Generator model. By providing Reddit user comments as ``knowledge,'' Audrey is able to make pop culture references to the topic, provided the comment contains no sensitive content.


\subsubsection{News Response Generator} \label{sec:newsgenerator}

Sourcing the latest news to customers who are interested in knowing and chatting about them is an important part of any social conversation.  Our main source for our news generator is the Washington Post.  Using the API provided by the Alexa Prize Team, we were able to access articles from the Washington Post.  The API provided basic information about the article, such as the author, topic, and the body of the text.  Through basic pre-processing and using the TextRank method for summarization \cite{mihalcea-tarau-2004-textrank}, we were able to split the body of the text into bite size information for a conversation. We used a template based approach to discuss recent news items. This is in similar fashion to the Trending News Conversation Generator, but the conversation is more reliably generated and fact-based by having consistent access to the content of the news articles that are fed into a template-based approach.

\begin{table}[ht!]
\begin{tabular}{l|p{12cm}}
Audrey & I have recent metro transit police news! Would you like to hear this one about Previous hoaxes from “Kid Cole” landed him on MTV’s show? \\
 \cellcolor{gray!5}User &  \cellcolor{gray!5}Sure. \\
Audrey & Let me summarize for you. It basically says the following. Jerez Nehemiah Stone-Coleman, 21, a.k.a. Kidd Cole, pleaded guilty Dec. 18 to one count of making threats involving explosive materials in a plea deal, after admitting to making 13 terrorist-related threats among what prosecutors said were more than 300 calls to 911 over a six-month period before his May 27 arrest. Shall I go on? \\
 \cellcolor{gray!5}User &  \cellcolor{gray!5}Yes. \\
Audrey & Mehta said he "thought long and hard" about Stone-Coleman's punishment, saying it might not be the "best outcome" from the defendant's perspective but was appropriate under the law. \\
 \cellcolor{gray!5}User &  \cellcolor{gray!5}That's interesting, I hadn't heard about it. \\
\end{tabular}
\caption{\label{tab:table-name2}Conversation example of the news response generator. To respect customers’ privacy, this is not a real customer conversation.}\
\end{table}

The news response generator is able to discuss factually-based information about Washington Post articles, providing a stable conversation about a variety of recent topics.

\subsubsection{Topical Response Generator}
\label{sec:topical-response}

The topical response generator is a dynamic retrieval-based generator that can engage customers in a particular topic before switching to another topic depending on the customers' interest levels. We analyzed interests from Alexa Prize Social Bot customer feedback and hand picked non-mainstream topics such as arts, gaming, and Pokemon, etc.  These highly tailored topical generators showed great engagement with the customers. Below is a list of topics that we implemented in chronological order. 

\begin{itemize}
    \item \textbf{Fitness} - The Fitness topic module talks about different forms of exercise, such as cardio, strength training, yoga, and flexibility routines. Our responses were constructed in an encouraging tone and in a customer-friendly manner. We also included some responses to keep the conversation flowing even if the customer is not interested in fitness. Similar to mainstream models like movies, music, and books, the hierarchy was created to account for all possible customer responses.
    
    \item \textbf{Season} - The Season topic module talks about different activities for each season, such as going to the beach and hiking in the summer. The hierarchy begins with the four seasons in general, and then continues to the second level, which contains details about the specific activities that are popular in each season. From these starters, we were able to understand what a particular customer likes to talk about related to their choice of season, which helped us to provide appropriate follow-ups to the customers' responses. An example of a conversation related to the Season module is shown in Table \ref{tab:topicalchat-example}.
    
    \item \textbf{Food} - The Food topic module talks about food related topics varying from exotic international cuisine to the top ten ice cream flavors. There were a lot of possibilities with this module, because food has a wide range of varieties based on the taste, the process of making, and, most importantly, based on customers' own preferences, which gave us lots of dimensions to talk about. To tackle this, we first created a decision tree to predict how the customer will respond to certain questions and to maintain and deepen the conversation within context, then we constructed our responses based on the customer's responses.
    
    \item \textbf{Weather} - The Weather topic module talks about a customer's favorite weather and outdoor activities related to that weather. The customers prefer different activities in different weather environments, so we start the conversation with weather in general and go into detail when we grasp the customers' favorite weather or the one they are interested in talking about. We also added follow-up questions at the end of our response to continue the conversation.
   
    \item \textbf{Game} - The Game topic module talks about flagship games in different genres, such as League of Legends (MOBA), Overwatch (FPS), The Witcher 3 (RPG), and Goat Simulator (SIM). The hierarchical structure goes from the different genres to the specific games customers are interested in. This module also contains appropriate responses for customers that are not interested in gaming by routing to other topics such as fitness and movies to continue the flow of the conversation.
   
    \item \textbf{Pets} - The Pets topic module talks about the common household pets, such as dogs, cats, and fish, and their favorite toys. With the wide range of pet choices, this module was complex and filled with responses that intertwined the subtopics together. To ease the flow of this topic module, we started with a question that asks whether the customer has a pet or not. If yes, then the conversation continued to talk about their pet. In the other case, we continued with their preferences regarding pets.
   
    \item \textbf{Art} - The Art topic module talks about  different forms of art, such as painting and sculpting. The hierarchy starts with two general forms of art, such as visual art and performing arts. The decision tree then continues to outline the specific art forms within each general category. While responding to their favorite art forms, we continue the conversation with discussions such as the art marking process and different art mediums.
   
    \item \textbf{Technology} - The Technology module talks about ubiquitous technologies, such as smartphones, game consoles, and the internet. With a broad sense of what the customer's favorite technology is, we then continue with responses relating to their functionalities and the common everyday activities such as social media and productive work. The module also branches to other possible related topics such as art and film to have more variety.
                                    
    \item \textbf{Sports} - The Sports topic module talks about different sports including football, basketball, and swimming. Each sport leads to specific questions that learn more about the customer's interests regarding the sport. 
    In case the customer is interested in a sport that was not in our database, we are still able to continue the conversation by referring to the sport the customer mentioned and asking meaningful questions regarding their interests.
   
    \item \textbf{Science} - The Science topic module talks about biology, physics, and chemistry. For the customers interested in science, we have branches in the decision tree that lead to discussion about the details of each subject. On the other hand, if the customer wasn't interested in science specifically, we were able to route them to other topics such as games and food to keep the conversation engaging.
   
    \item \textbf{Pokemon} - The Pokemon topic module talks about twenty types of Pokemon, including but not limited to fire type, water type, and air type. From then on we delve deeper into details about each type of Pokemon along with the skilled Pokemon trainers. This module also includes introductions to other topic modules to ease the transitions between different modules.
    
    \item \textbf{Anime} - The Anime topic module talks about seventeen different popular anime. From there we ask the customer's opinions on different parts of the anime, such as fight scenes or the opening. If the customer wasn't interested in anime, we could route them to another topic such as movies.

\end{itemize}

\begin{table}[ht!]
    \begin{tabular}{l|p{12cm}}
    Audrey & I love seeing how the world changes from season to season. What is your favorite season? \\
     \cellcolor{gray!5}User &  \cellcolor{gray!5}Spring. \\
    Audrey & Ah Spring makes me happy! The world is filled with color and excitement! What do you like most about Spring? \\
     \cellcolor{gray!5}User &  \cellcolor{gray!5}The warm weather. \\
    Audrey & The warm weather is definitely a welcome change after the bitter winter. Have you watched spring cherry blossoms?  \\
     \cellcolor{gray!5}User &  \cellcolor{gray!5}Yes I have! \\
    \end{tabular}
    \caption{\label{tab:topicalchat-example}An example conversation from the Season topic module. To respect customers’ privacy, this is not a real customer conversation.}\
    \vspace{-4mm}
\end{table}

The core of our topical response generator module is a custom, flexible framework that retrieves sentences from a curated database. The generator uses the raw customer utterance, noun entities, and customer's previous turn history to retrieve the best responses from our database. The database is a list of nested dictionaries that stores pairs of expected customer prompts and Audrey's template-based responses. The expected customer prompts are phrases that we expect customer to utter which we then pair up with the most appropriate bot response based on templates within each topic module.


When generating a response, the customer's     turn history is used to index into the corresponding location in our database, limiting our search from thousands of response templates to just tens of responses. Then, we select the optimal response by matching the customer utterance to the most similar expected customer prompts via a weighted combination of keywords search and cosine similarity matching between sentence vectors given by Spacy's pre-trained word embedding\cite{honnibal-johnson:2015:EMNLP}. Once an optimal response template is found, we fill the template with noun phrases, verbs, or adjectives extracted from the customer utterance. If there is no ideal template based response to a particular customer utterance, we use the neural response generator as a backup. 

\subsubsection{Hierarchical Topical Generator} \label{sec:hiearchy}


In order to design a unique experience for the customers who engage with Audrey, we built a Hierarchical Topical Generator module to have flexibility, opinion and engagement. Our model was based on the data provided by the Amazon's Evi Knowledge Graph from Section \ref{sec:kg}. We defined a set of attributes for each topic module using the entity. For example, given a movie title from a customer utterance, attributes like its actors, directors, plot or other related movies' information can be extracted through Evi. The attributes were defined so that the information regarding these attributes could be maintained in all conversation turns within that particular topic module, while topic modules are initiated based on the extracted entity (like movie title here will initiate the Movies Module). On top of that, we also define a hierarchical structure for all of these attributes within each topic module. In each turn, the generator selects a topic attribute and generates a relevant responseby accessing related information from the knowledge graph. When enough context information is not present, we follow the defined hierarchy to select the attributes.\par

We flexibly switch between attributes through an interplay of questions and opinions when enough context about customer preference is available and the switches between topics are handled by the dialogue policy. The design of our hierarchical topical generator allows us to have long, in-depth and engaging conversations with customers about these topics.


Audrey's static knowledge comes from a mixture of primarily domain-specific datasets. For movie knowledge, The Movie Database\footnote{https://www.themoviedb.org/} and The Open Movie Database\footnote{https://www.omdbapi.com/} were fused together into a Amazon DynamoDB table\footnote{https://aws.amazon.com/dynamodb/} to provide metadata information about popular movies, such as its title, abridged plot summary, ratings, and actors.

Other static data, less focused on knowledge, included several years of Reddit comments. Using a rigorous word blacklist, as well as a subreddit blacklist for sensitive content, Reddit comments were filtered and indexed into Amazon's ElasticSearch \footnote{https://aws.amazon.com/elasticsearch-service/}. This allowed for keywords and phrases to be queried and quickly returned with relevant Reddit comments containing that keyword or keyphrase. Additionally, items could be queried by subreddit, allowing for versatile pool of opinionated comments on nearly any topic, such as movie opinions that reference a certain movie title in /r/movies. We observed that opinion based response generation leads to better engagement with customers than stating facts or summaries.



\subsubsection{TopicalGPT Response Generator} 
\label{sec:TopicalGPT}

In the above sections, we designed diverse response generators to deal with different topics uttered frequently in the conversations.  However, general conversation without any specific topic or chit-chat also constitutes an important part of the conversation while talking to a chatbot. Customers may talk about a random topic with Audrey at the very start or maybe after several turns into the conversation which adds more uncertainty to the conversation and increases the difficulty of designing a dialog system which can respond appropriately. \par

In order to deal with open domain chat, we took advantage of one of the state-of-the-art natural language generation models, GPT-2 \cite{radford2019language}, a large scale unsupervised language model. GPT-2 has shown excellent performance and large capacity to generalize in many natural language generation tasks, which led us to integrate the GPT-2 model directly without any modifications. Specifically, we treat the last few turns of customers' conversation as the context and input it to GPT-2 the model which then generates the relative response based on it. For training purposes, we used the Topical-Chat dataset released by Amazon \cite{Gopalakrishnan2019} to fine-tune the model. This dataset includes more than 235,000 utterances and generalizes the pattern of chit-chat or open domain conversation very comprehensively. We also proposed a multi-task objective function for training the model: on one hand, we try to minimize the perplexity of the output sentence and on the other hand, we add an extra classification task---given the context, we provide the real response and a bunch of unrelated sentences, and let the model to select the correct one. We believe these tasks drives the model to generate output both with high fluency and accuracy. To show the advantage of GPT-2 model, we ran an experiment to compare the performances between GPT and GPT-2;  the result is shown in Table \ref{tab:gpt} and confirm the advantage of GPT-2 over GPT in our setting. Examples of generated responses from our model are shown in Table \ref{tab:neuralGenerationExample}.

\begin{table}[t]

  \centering
  \begin{tabular}{lll}
    \toprule
    \cmidrule{1-3}
    Models     & Negative log likelihood & Perplexity  \\
    \midrule
    GPT & 2.904 & 18.25    \\
    GPT-2 &2.891 &18.02    \\
    \bottomrule
  \end{tabular}
  \setlength{\abovecaptionskip}{2pt}
  \caption{Performance Comparison Between GPT and GPT-2}
  \label{tab:gpt}  
\end{table}

\begin{table}[ht!]
    \begin{tabular}{l|p{12cm}}
    \cellcolor{gray!5}User & \cellcolor{gray!5}Do you like dogs or cats? \\
    Audrey & I'm a dog person, but my cats love me. \\
    \end{tabular}
    \caption{\label{tab:neuralGenerationExample}Topic variety with the TopicalGPT module. To respect customers’ privacy, this is not a real customer conversation.}\
    \vspace{-4mm}
\end{table}


\subsubsection{Empathetic Response Generator}
\label{sec:empathetic}
Acknowledging customers emotions and providing relevant responses suitable for each kind of emotion is a key element in any social conversation. Based on this idea, we decided to use the classified emotion determined from a customer's past context as a cue to our generator model to generate an empathetic response. The generator would be evoked by the Dialog Policy when the emotion would be classified as either \textit{happy} or \textit{angry}.

Similar to the TopicalGPT response generator, we use GPT-2 model as the backbone of this generator model. We fine-tuned GPT-2 model using Empathetic Dialogues dataset which is also used in creating the emotion detection classifier.

\begin{table}[ht!]
    \begin{tabular}{l|p{12cm}}
    \cellcolor{gray!5}User & \cellcolor{gray!5}I am feeling sad today, I did poorly on my exam. \\
    Audrey & That's okay, just keep working hard and you'll be fine! \\
    \end{tabular}
    \setlength{\abovecaptionskip}{2pt}
    \caption{\label{tab:empthateticGenerationExample}Empathetic Generator variety with the Empathetic Response Generator. To respect customers’ privacy, this is not a real customer conversation.}\
    \vspace{-4mm}
\end{table}

\subsubsection{Amazon Neural Response Generator} \label{sec:amazonneural}


The Amazon Neural Response Generator was provided as a service starting in the quarter-finals interaction period.  The model was trained based on the transfer learning approach by HuggingFace\cite{wolf2019transfertransfo}.  The generator was used to handle out-of-domain responses as well as generating responses for the trending news conversation module.


\subsection{Prosody}

Our prosody generation utilizes Amazon Alexa’s speech synthesis system in order to give Audrey her own unique voice. We used the Amazon Speech Synthesis Markup Language(SSML) format to enhance our templates by dynamically adding tags to our final output response. For example, whenever we encountered a question mark or exclamation mark in the generated response, we inserted emotion and pitch tags in order to provide the inflection to indicate more inquisitive or curious responses. After testing different samples with various SSML tags and hierarchies for introducing prosody, we decided to use excited emotion ("<amazon:emotion name='excited' intensity='low'>") for longer responses and ("<prosody pitch='high'>") for shorter responses. For longer responses generated by our bot, we also experimented with increasing the speaking rate, bringing it closer to normal speaking patterns.

\section{Results and Analysis}
During the semi-finals interaction period from March 20\textsuperscript{th} to April 29\textsuperscript{th} we performed a series of experiments to quantify the impact of different parts of Audrey's pipeline on conversational ratings and duration. We describe insights from descriptive analyses done on these conversations in Section \ref{sec:analysis} and comparative experiments on conversation quality done through systematically adjusting Audrey's components in Section \ref{sec:experiments}.


\subsection{Analysis} \label{sec:analysis}

The following five analyses in Sections \ref{sec:starteranalysis}--\ref{sec:topicanalysis}  reflect Audrey's performance during the semi-final interaction period during March 20\textsuperscript{th} to April 29\textsuperscript{th}.


\begin{figure}[h!]
  \centering
  \includegraphics[width=0.5\textwidth]{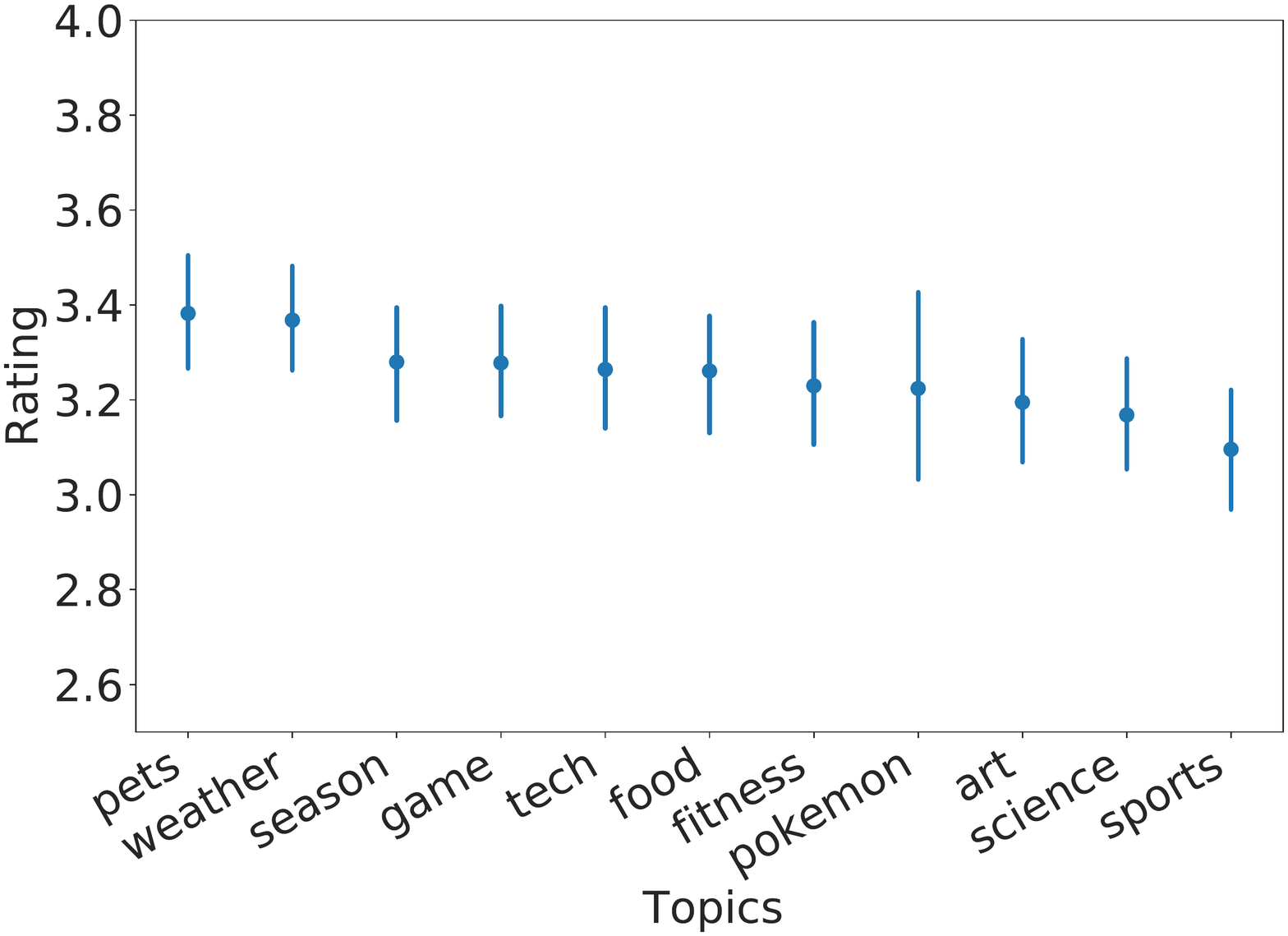}
  \caption{Overview of ratings vs conversation starters during the semi-finals interaction period. When we started our conversations with pets and weather topics, we had highest initial engagement.(Note that we changed the y-axis from 2.5-4.0 for easier visualization.)}
  \label{fig:converation-starter}
\end{figure}

\subsubsection{Which conversation starters gave better ratings?} \label{sec:starteranalysis}

Our conversations starters aim to initiate the conversation on a high note and quickly engage the customer in a topic of their interest that Audrey is also able to chat about.  Figure \ref{fig:converation-starter} shows the resulting conversation score based on which started was used to initiate the dialog. While conversations can go many directions after the starter, these results indicate that customers consistently preferred to start the conversation with ``light'' conversation fare, e.g., about the weather or their pets, rather than focusing directly on starters that are more domain-interests. Our results highlight the importance of small talk \cite{coupland2014small}  as an avenue for drawing new customers into a conversation.



\subsubsection{What is the relationship between conversation length and rating?}
\label{sec:time-and-rating}
\begin{figure}[h!]
\centering
    \includegraphics[width=0.45\textwidth]{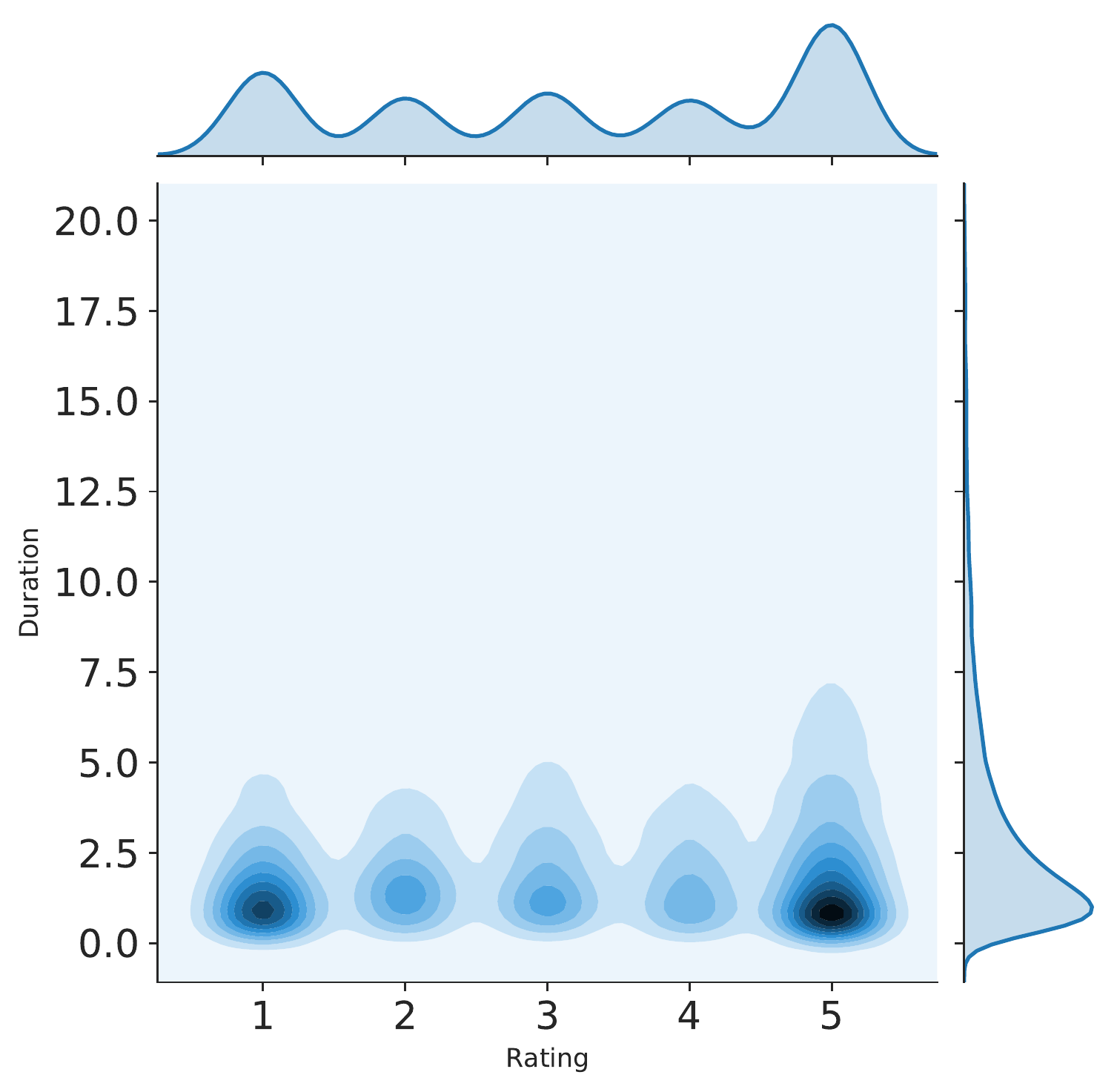}
    \caption{Distribution of ratings and duration. Conversation Rating and Duration were weakly correlated (Pearson r=0.13) suggesting that customers willing to talk for a long duration with Audrey are not more satisfied with the quality than those only willing to talk for a shorter duration.}
    \label{fig:duration1}
\end{figure}

Customers who keep talking to Audrey are able to experience a wider breadth of topics. Ten percent of our conversations lasted longer than 7 minutes and 25 seconds. It is intuitive that longer conversation correlated with higher ratings, but Figure \ref{fig:duration1} shows that duration is only weakly correlated with rating.


\subsubsection{How are ratings different for new and returning customers?} \label{sec:newreturning_analysis}

The Alexa Prize Grand Challenge platform allows customers to return, which provides an opportunity for Audrey to use information learned about them during their prior conversations (e.g., interests in particular movies or sports) to engage with them. However, returning customers also present a challenge to our generators, as customers may have heard some of the content before (e.g., retreading the previous conversation). Therefore, we tested whether the conversation score was higher for returning versus new customers.\footnote{Note that because the customer identifiers we receive reset periodically, we are unable to track all return visits and some first visits by a new customer may actually be returning. Additionally, due to the blinded nature of customer interactions with the Alexa Prize experience, many first time visitors may have previously interacted with other socialbots.}

\begin{figure}[tb!]
\centering
    \includegraphics[width=0.45\textwidth]{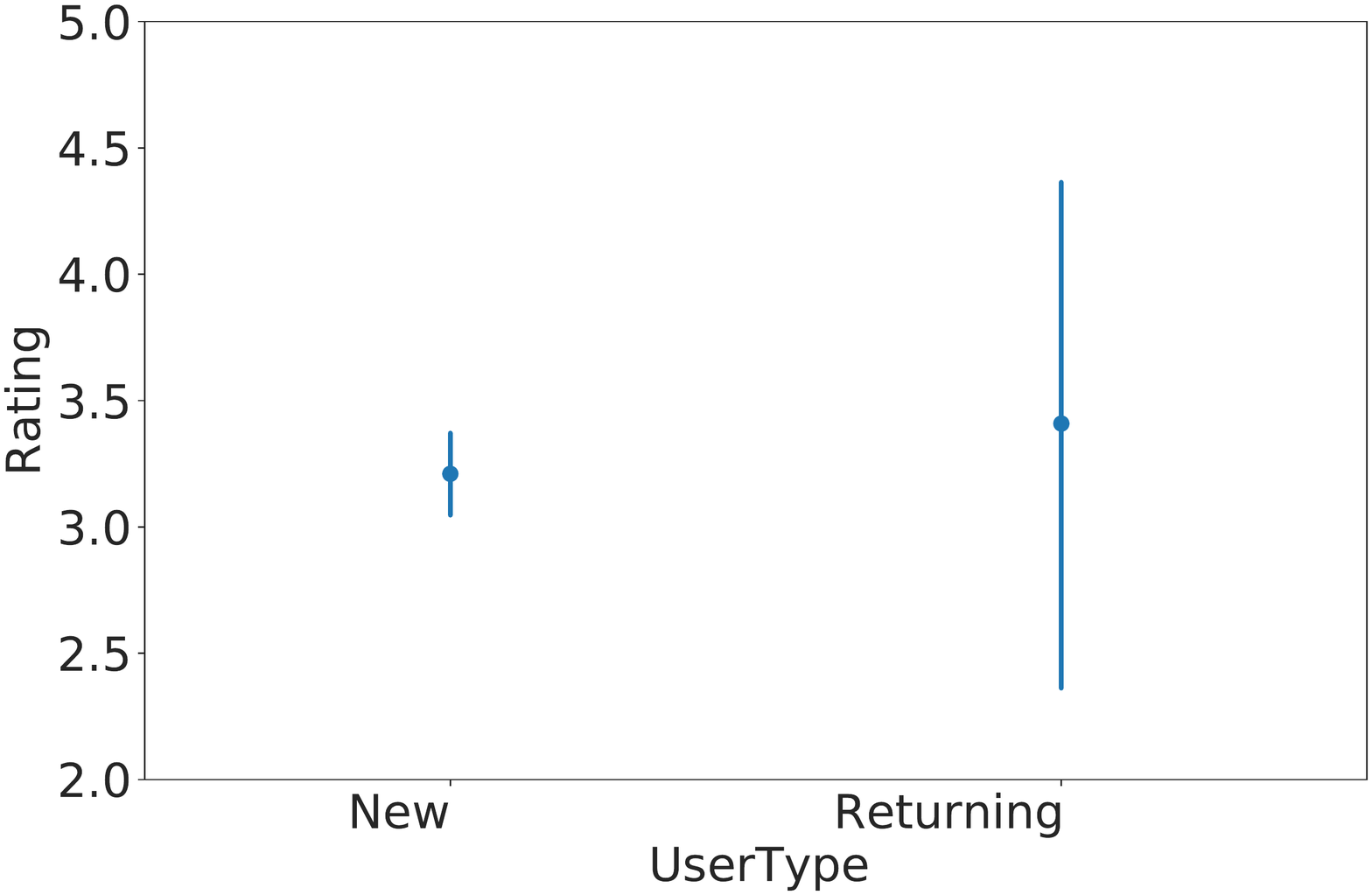}
    \caption{Point Plot based on new and returning customer comparison.  Median ratings for returning customers were higher than the new customers.}
    \label{fig:ratingvsusertype}
\end{figure}

Figure \ref{fig:ratingvsusertype} shows a plot of the mean and the spread of ratings for new and returning customers. New customers gave an average rating of 3.21, and returning customers gave an average rating of 3.40. While returning customers gave a higher average rating, the spread is much larger compared to new customers.

\subsubsection{How do different topic modules and generators affect the engagement of the customer?}

\begin{figure}[th]
  \centering
  \subfloat[Generators vs Average number of  turns]{\includegraphics[width=0.5\textwidth]{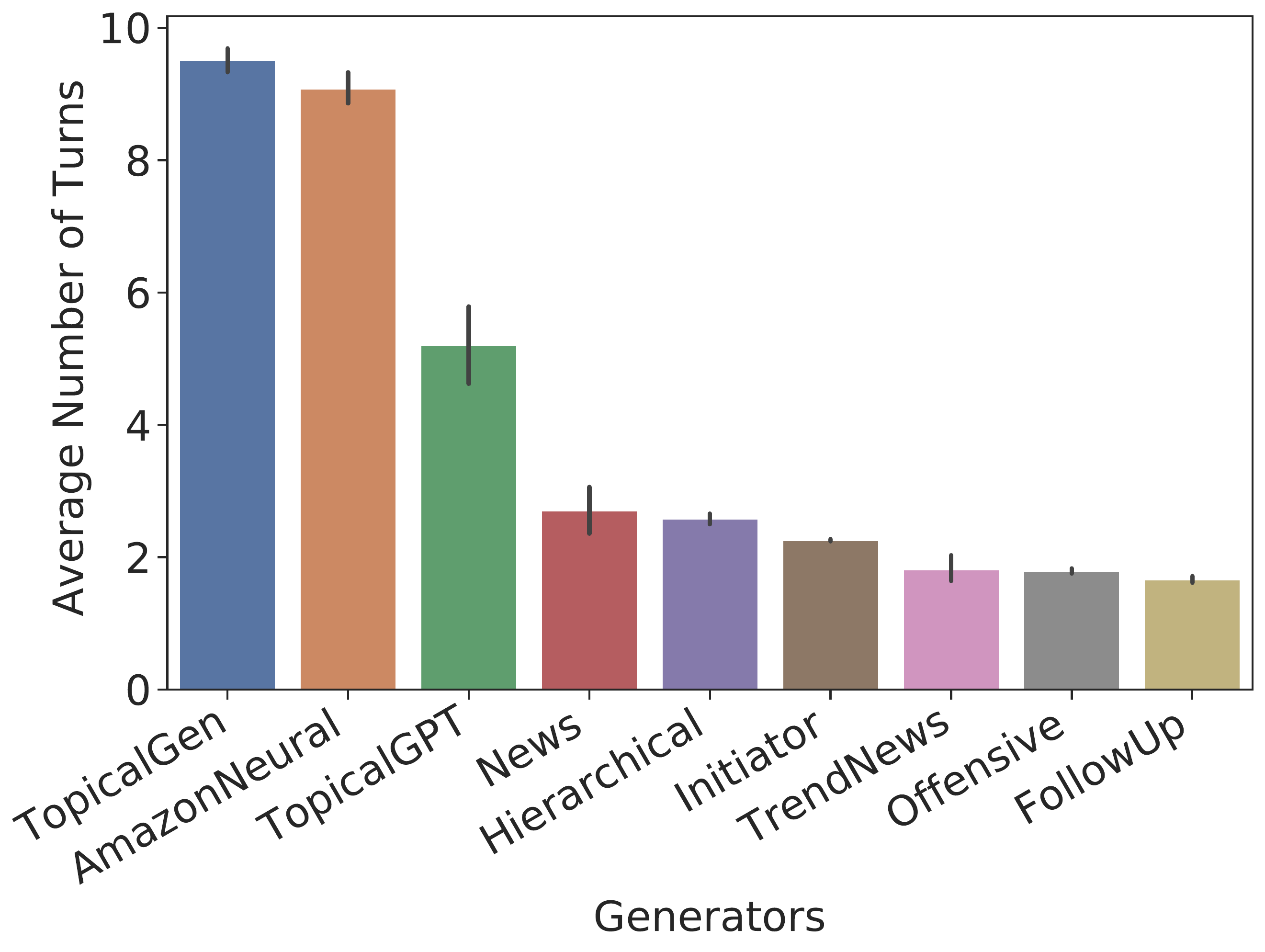}\label{fig:d1}}
  \hfill
  \subfloat[Topics vs Average number of  turns]{\includegraphics[width=0.5\textwidth]{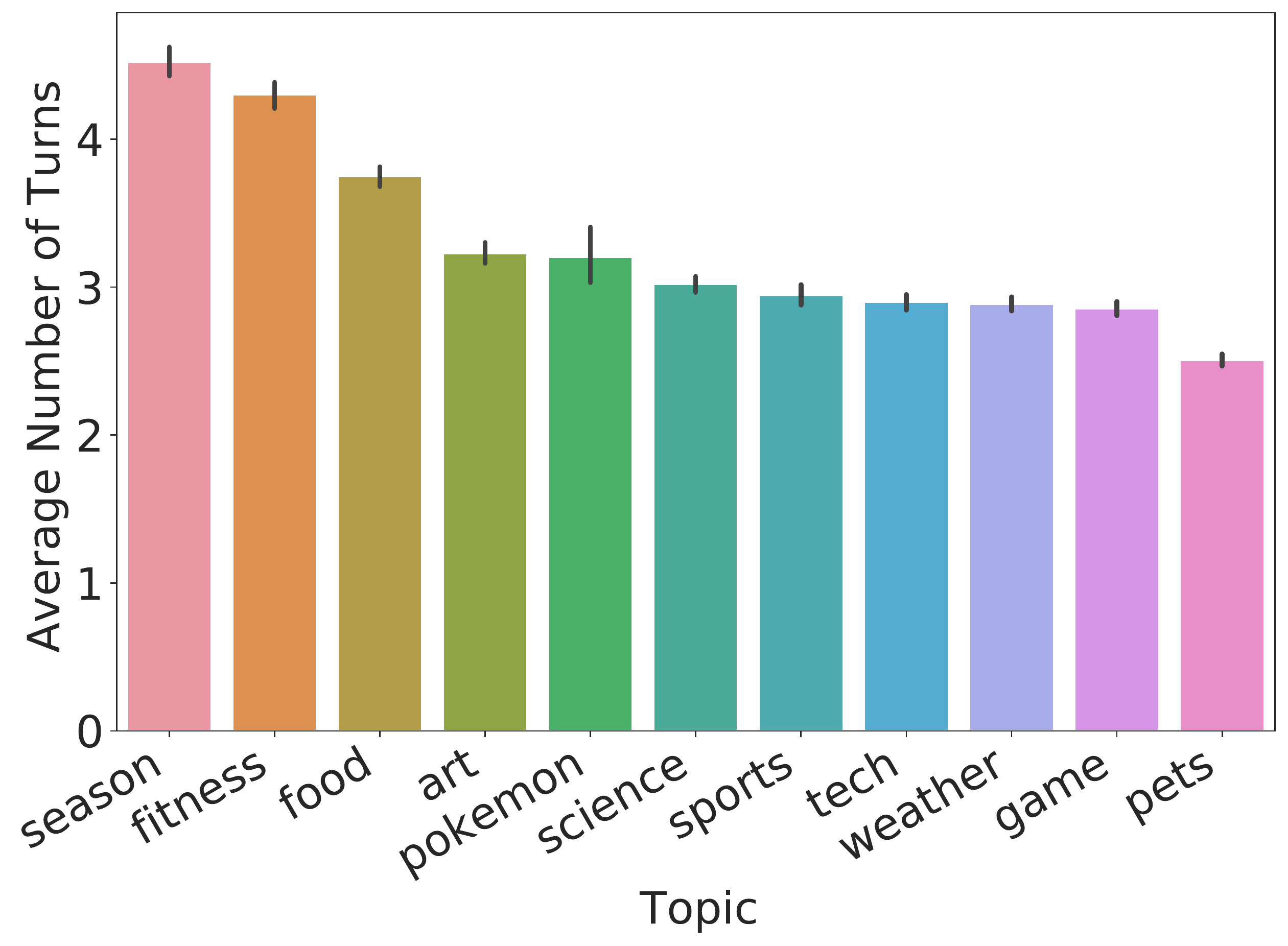}\label{fig:d2}}
  \caption{Overview of average number of turns vs generators and topics. a) The topical chats and neural generators engage customers the most. b) Among the topical chats, season and fitness topics get the longest conversations.}
  \label{fig:duration}
\end{figure}

Audrey uses multiple modules for generating speech (\Sref{sec:topical-response}) guided by conversation policy goals (\Sref{sec:rl} and \Sref{sec:pum}). These generators each can have different effects on the engagement of the customer, based on their interests. Here, we measure engagement through the number of responses made by a single generator---i.e., did the customer talk to this part of Audrey more. 
%
%
Figure \ref{fig:d1} shows that customers consistently engaged more  with the neural response generators than with other template or hybrid generators. 
The most-utilized generator, Topical Response, contains multiple topics, each could have different levels of engagement. Therefore, we examined the average number of turns for each topic, shown in Figure \ref{fig:d2}.
Among topics, season and fitness led to more conversation, with over a full turn more dialog. Customers engage with the majority of topics for at least three turns on average with the exception of pets which was slightly lower (2.50). We speculate that the pets topic results in fewer turns due to  conversational redirection from people who don't have pets.

\subsubsection{How do different topic modules and response intents affect the overall ratings?} \label{sec:topicanalysis}

Because Audrey's responses are generated by different initiators and topic modules, many factors can affect overall rating. In order to pinpoint the modules that are positively affecting our performance, we construct an OLS regression where we treat each conversation as an observation and include independent variables for (i) the number of turns from each generator, (ii) the particular conversation starter,  (iii) generator expressions like asking for a customer to repeat themselves, (iv) sensitive questions uttered by the customer, and (v) the duration of the conversation in minutes. By using the customer rating as the dependent variable, this model let's us quantify the impact of each component.

The regression results, shown in Table \ref{fig:olsRegression}, contain four main findings.
First, the largest significant effect on the conversation comes from longer conversations with the topical generators. Surprisingly, despite Pets having the shortest overall conversation (cf. Figure~\ref{fig:duration}b) discussions on Pets has the largest positive effect on score among all topics. Other strong contributors were more narrow topics such as Art, Fitness, Food and Games.
Second, our knowledge-rich hierarchical response generators (built on top of EVI) did not significantly improve conversations, despite their ability go deeper into a topic based on domain knowledge.
Third, we observe a statistically-significant negative effect for when Audrey uses Topical GPT directly to generate a response (not as a part of a strategy) based on what the customer has said. As this fallback strategy is evoked when Audrey cannot determine how to best proceed, this coefficient suggests that (i) a better strategy for handling non-sequitur comments by customers can improve conversation quality and that (ii) neural generators are not yet suitable to fully dynamically generate the conversation over long periods.
Fourth, building upon the analysis of the initiators (\Sref{sec:initiators}), when controlling for all other factors, surprisingly, the way the conversation is initiated has no significant impact on the resulting score---though the most positive initiator, talking about the Seasons, approaches significance with a large positive coefficient.

\begin{table}[!htbp] \centering
    \begin{tabular}{@{\extracolsep{5pt}}lSc}
    & $\beta$ & Standard Error \
    \cr \cline{1-2}
    \hline \\[-1.8ex]
     \cellcolor{gray!10}\textit{Intercept} & 2.948$^{***}$ & 0.104\\
     \cellcolor{gray!25}Duration (minutes) & 0.001$^{***}$ & 0.000\\
     \cellcolor{red!15}Conversation Starter: None & 0.095$^{}$ & 0.105\\
     \cellcolor{red!15}Conversation Starter: Art & 0.079$^{}$ & 0.118\\
     \cellcolor{red!15}Conversation Starter: Fitness & 0.045$^{}$ & 0.110 \\
     \cellcolor{red!15}Conversation Starter: Food & 0.067$^{}$ & 0.110\\
     \cellcolor{red!15}Conversation Starter: Games & 0.071$^{}$ & 0.117\\
     \cellcolor{red!15}Conversation Starter: Pets & 0.071$^{}$ & 0.117\\
     \cellcolor{red!15}Conversation Starter: Pokemon & 0.132$^{}$ & 0.144\\
     \cellcolor{red!15}Conversation Starter: Science & 0.011$^{}$ & 0.120\\
     \cellcolor{red!15}Conversation Starter: Seasons & 0.184$^{*}$ & 0.110\\
     \cellcolor{red!15}Conversation Starter: Sports & -0.073$^{}$ & 0.123\\
     \cellcolor{red!15}Conversation Starter: Tech & 0.043$^{}$ & 0.120\\
     \cellcolor{red!15}Conversation Starter: Weather & 0.170$^{}$ & 0.117\\
     \cellcolor{magenta!35}Hierarchical Response: Books &  0.022$^{}$ & 0.017\\
     \cellcolor{magenta!35}Hierarchical Response: Movies & 0.028$^{*}$ & 0.016\\
     \cellcolor{magenta!35}Hierarchical Response: Music & 0.013$^{}$ & 0.022\\
     \cellcolor{blue!25}Topical Response: Art & 0.031$^{**}$ & 0.013\\
     \cellcolor{blue!25}Topical Response: Fitness & 0.039$^{***}$ & 0.007\\
     \cellcolor{blue!25}Topical Response: Food & 0.034$^{***}$ & 0.009\\
     \cellcolor{blue!25}Topical Response: Games& 0.043$^{***}$ & 0.014\\
     \cellcolor{blue!25}Topical Response: Pets & 0.055$^{***}$ & 0.016\\
     \cellcolor{blue!25}Topical Response: Science & 0.027$^{**}$ & 0.014\\
     \cellcolor{blue!25}Topical Response: Season & 0.028$^{***}$ & 0.006\\
     \cellcolor{blue!25}Topical Response: Sports & 0.016$^{}$ & 0.015\\
     \cellcolor{blue!25}Topical Response: Tech & 0.028$^{**}$ & 0.014\\
     \cellcolor{blue!25}Topical Response: Weather & 0.018$^{}$ & 0.014\\
     \cellcolor{cyan!25}Amazon Neural Generator & -0.006$^{***}$ & 0.002\\
     \cellcolor{green!25}Follow Up Questions & -0.011$^{}$ & 0.013\\
     \cellcolor{orange!25}Goodbye Response & 0.042$^{*}$ & 0.024\\
     \cellcolor{red!25}Response to Offensive Comments & -0.048$^{***}$ & 0.014\\
     \cellcolor{magenta!45}Repeat Response & 0.047$^{}$ & 0.032\\
    \hline\hline \\[-1.8ex]
     Observations & { 16,404 }\\ 
     R$^2$ & 0.033 \\
    \hline
    \hline \\[-1.8ex]
    \textit{Note:} & \multicolumn{1}{r}{$^{*}$p$<$0.1; $^{**}$p$<$0.05; $^{***}$p$<$0.01} \\
    \end{tabular}
    \caption{OLS regression result for different intents that could effect the overall rating. }
    \label{fig:olsRegression}

\end{table}

\subsection{Experiments} 
\label{sec:experiments}

Throughout the quarterfinals and semifinals periods, multiple A/B tests were done on the platform to quantify the effect of specific components. In the following tables, the average conversation duration are adjusted to exclude outliers.

\subsubsection{Intimacy Experiment for transitions} \label{sec:exp-intimacy}

Our question intimacy estimation modules enable Audrey to potentially move from more casual to deeper questions during conversation. To test whether systematically increasing question intimacy leads to higher-ranked conversations, we randomly selected conversations to engage in one of two strategies: (1) asking conversational questions in increasing order of intimacy or (2) asking the same questions in a random order, using the 76 questions from Section \ref{sec:followup}. This experiment was conducted over many conversations from March 28th to April 6th.

We hypothesized that by initially picking less intimate conversational questions for transitions in the beginning and gradually increasing the intimacy level of the question, customers would have a better feeling of the conversation that would translate into higher ratings.  However, the results in Table \ref{tab:intimacy-exp} showed a statistically-significant drop in conversational score, when questions were asked in increasing order. We speculate that while the questions were intriguing, they were not always asked at times that lead to strong conversational coherence, which negated the effect of question ordering.

\begin{table}[h!]
  \centering
  \begin{tabular}{lcccc}
    \toprule
    \cmidrule{1-3}
    Variant     & \makecell{Avg. Feedback \\ Rating} & Feedback CI & \makecell{Avg. Conversation \\ Duration (seconds)} & \makecell{Avg. Conversation \\ Duration CI} \\
    \midrule
    Random order & 3.321 & (3.287, 3.355) &  223.22 & (216.097, 230.343) \\
    Increasing intimacy  & 3.201 & (3.167, 3.235) &  196.33 & (190.076, 202.584) \\
    \bottomrule
  \end{tabular}
  \setlength{\abovecaptionskip}{2pt}
  \caption{Mean conversation ratings and duration when transitioning between topics using questions ordered by increasing intimacy versus the same questions asked in a random order. Intimacy had no effect on customers' preferences for the conversation, but increasing intimacy led to a statistically-significant drop in conversation duration. }
  \label{tab:intimacy-exp}
  \vspace{-6mm}
\end{table}

\subsubsection{Trending News Conversation Generator}
\label{sec:tncg-exp}

Our Trending News Conversation Generator (TNCG) allows Audrey to inject interesting discussion points during the middle of a conversation when the Dialog Policy manager (\Sref{sec:dp}) has identified that the customer's focus has drifted, potentially sparking more engagement. However, such transitions could seem out of place and jar the customer, especially with reliance on Topical GPT to generate subsequent conversation on recent news. Here, we performed an A/B experiment to test the effect of introducing recent news in a conversation by randomly varying whether the TNCG (\Sref{sec:dynamic} or Follow-up Generator (\Sref{sec:followup}) was used to transition between topics. The experiment was run during April 21st to April 23rd.
Our results in Table \ref{tab:dynamic-gen-exp} showed a statistically-significant increase in the average conversation duration with the TNCG. However, the generator had no significant effect on rating, a discrepancy that agrees with our findings in \Sref{sec:time-and-rating} that the two are weakly correlated. We speculate that TNCG generally increased conversation duration based on the interest of recent news over templated topic changers, but that its under-preparedness to discuss COVID-19 caused inconsistency with the final ratings. We observed a general lack of interest in discussing coronavirus news, which encompasses nearly our entire pool of content, despite selecting for positive-only news.

\begin{table}[h!]
  \centering
  \resizebox{0.98\textwidth}{!}{ 
  \begin{tabular}{l cccc}
    \toprule
    \cmidrule{1-3}
    Variant     & \makecell{Avg. Feedback \\ Rating} & Feedback CI & \makecell{Avg. Conversation \\ Duration (seconds)} & \makecell{Avg. Conversation \\ Duration CI} \\
    \midrule
    \makecell{Trending News  Gen.} & 3.276 & (3.217, 3.330) & 224.14 & (212.069, 236.211) \\
    \makecell{Followup  Generator} & 3.251 & (3.194, 3.308) & 151.00 & (109.665, 132.3348) \\
    \bottomrule
  \end{tabular}
  }
  \setlength{\abovecaptionskip}{2pt}
  \caption{Experiment on Trending News Conversation Generator and Follow-up Generator.}
  \label{tab:dynamic-gen-exp}
\end{table}

\subsubsection{Experiment on Providing Context to the Neural Generators} \label{sec:neuralexperiments}

The neural response generator (NRG; \Sref{sec:amazonneural}) conditions its output based on prior context. More context can potentially provide richer information to craft a conversational arc matching the current trajectory; however, selecting prior turns that cover multiple topics may muddy the output. Therefore, we conducted an A/B experiment to test the effect of different types of context provided to the NRG. The first condition uses all 
prior turns on the current topic as context, while the second always selects the 5 previous turns as context, regardless of their topic.


The experiment ran March 20th to March 28th with results shown in Table \ref{tab:neural-generator-exp}. 
Including only on-topic context for the NRG attained a mean rating of 3.308 compared against the 3.196 mean rating for using a fixed-size context. We observed the only on-topic context to be statistically different from fixed sized content, both in terms of higher average feedback ratings and lower average conversation duration. The inverted trends between ratings and conversation duration is unexpected, but this again corroborates with the findings in \Sref{sec:time-and-rating} that suggest only a weak correlation between the two. Nevertheless, we view this as a useful guide for the importance of conditioning on a more topically-narrow context to improve response quality and coherence.

\begin{table}[hbt!]
  \centering
  \resizebox{0.98\textwidth}{!}{ 
  \begin{tabular}{l cccc}
    \toprule
    \cmidrule{1-3}
    Variant     & \makecell{Avg. Feedback \\ Rating} & Feedback CI & \makecell{Avg. Conversation \\ Duration (seconds)} & \makecell{Avg. Conversation \\ Duration CI} \\
    \midrule
    Only On-topic Context  & 3.308 & (3.287, 3.329) & 208.06 & (204.130, 211.990) \\
    Fixed-size Context  & 3.196 & (3.181, 3.211) & 242.73 & (240.122, 245.338) \\
    \bottomrule
  \end{tabular}
  }
  \setlength{\abovecaptionskip}{2pt}
  \caption{Ratings Change on Introducing Dynamic Context Length for GPT-2}
  \label{tab:neural-generator-exp}
  \vspace{-6mm}
\end{table}

\section{Conclusion}

We designed an open-domain social conversation system, Audrey, that achieved a cumulative average rating of 3.25 out of 5 in the the semi-finals.  Audrey was designed with the goals of engaging with customers on a personal level by adapting to their preferences, interests, and personality. To achieve these goals, we developed a large collection of NLP modules for language understanding (\Sref{sec:nlu}, dialog management (\Sref{sec:dp}), and response
generation (\Sref{sec:nlg}) that create a diverse conversational landscape intended to evoke delight. 
%


\section{Future Work}

Given time constraints from late deployment in semifinals, we were not able to perform rigorous experimental analysis on the innovative features released recently like the Personal Understanding Module (PUM) and Reinforcement Learning (RL) based Topic Selection and some new topical modules. We will aim to improve the Personal Understanding Module (PUM) so that it can better suggest topics based on customers' likes and dislikes and create an adaptive and unique conversation experience for them. Lastly, we also plan to improve our reinforcement learning model to build a better dialog policy for topic selection and conversation content planning.

\section{Acknowledgement}

We would like to acknowledge the help from Amazon Alexa Prize team and Amazon Engineers for their constant technical support, financial support and cooperation during this challenge.

\bibliography{references}
\bibliographystyle{plain}





\end{document}